%% file: main.tex
\documentclass[runningheads]{llncs}
\usepackage{floatrow}
\usepackage{graphicx}
\usepackage{amsmath}
\usepackage{amsfonts}
\usepackage{amssymb}
\usepackage{mathrsfs}
\usepackage{upgreek}
\usepackage{pifont}  
\usepackage{booktabs} 
\usepackage{graphicx} 
\usepackage{xcolor}

\usepackage{svg}
\usepackage[export]{adjustbox}
\usepackage{enumerate}
\usepackage{multirow}
\usepackage{textgreek}

\usepackage{hhline}
\usepackage{array}
\usepackage{tabularx} 
\usepackage{caption}  
\usepackage{floatrow} 
\usepackage{graphicx,subfigure}
\usepackage{tabu}
\usepackage{siunitx}

\usepackage{verbatim} 
\usepackage{caption} 
\usepackage{color}
\usepackage{url}
\usepackage{xcolor}
\definecolor{cvprblue}{rgb}{0.21,0.49,0.74}
\usepackage[pagebackref,breaklinks,colorlinks,citecolor=cvprblue]{hyperref}
\newcommand{\medtit}[1]{\medbreak\noindent\textbf{#1.}}     
\begin{document}
\title{\texttt{GeoContrastNet}: Contrastive Key-Value Edge Learning for Language-Agnostic Document Understanding}

\titlerunning{Key-Value Edge Learning for Language-Agnostic Document Understanding}
%
%
%

\author{Nil Biescas\inst{1}$^,$  \inst{2}\orcidID{0009-0001-3722-4329} \and
Carlos Boned \inst{1}$^,$  \inst{2}\orcidID{0009-0000-6041-0931} \and
Josep Llad\'{o}s\inst{1}$^,$ \inst{2}\orcidID{0000-0002-4533-4739} \and
Sanket Biswas\inst{1}$^,$ \inst{2} \thanks{Main Corresponding Author}\orcidID{0000-0001-6648-8270} 
}

\authorrunning{N.Biescas et al.}
\institute{Computer Vision Center, Catalonia, Spain \\
               \email{Nil.Biescas@autonoma.cat, \{cboned, josep, sbiswas\}@cvc.uab.es} 
               \and
               Computer Science Department \\ 
               Universitat Autònoma de Barcelona, Catalonia, Spain }

%
\maketitle              
\begin{abstract}
This paper presents \textbf{GeoContrastNet}, a \textit{language-agnostic} framwork to structured document understanding (DU) by integrating a contrastive learning objective with graph attention networks (GATs), emphasizing the significant role of geometric features. We propose a novel methodology that combines geometric edge features with visual features within an overall two-staged GAT-based framework, demonstrating promising results in both link prediction and semantic entity recognition performance. Our findings reveal that combining both geometric and visual features could match the capabilities of large DU models that rely heavily on Optical Character Recognition (OCR) features in terms of performance accuracy and efficiency. This approach underscores the critical importance of relational layout information between the named text entities in a semi-structured layout of a page. Specifically, our results highlight the model's proficiency in identifying key-value relationships within the FUNSD dataset for forms and also discovering the spatial relationships in table-structured layouts for RVLCDIP business invoices. Our code is accessible on this GitHub$^\dagger$.
\def\thefootnote{$\dagger$}\footnotetext{\url{https://github.com/NilBiescas/GeoContrastNet}}

\end{abstract}
\input{tex/intro}

\input{tex/related}
\input{tex/method}

\input{tex/experiments}

\input{tex/conclusion}

\section*{Acknowledgment}
Work co-supported by the Spanish projects PID2021-126808OB-I00 (GRAIL) and CNS2022-135947 (DOLORES), the Catalan project 2021 SGR 01559, the PhD Scholarship from AGAUR 2023 FI-3-00223 and the CVC Rosa Sensat Student Fellow. The Computer Vision Center is part of the CERCA Program/Generalitat de Catalunya.

\bibliographystyle{splncs04}
\bibliography{main}
\newpage

\end{document}

%% file: tex/intro.tex
\section{Introduction}



 
 Visual information extraction (VIE)~\cite{davis2019deep,davis2022end,gemelli2022doc2graph,FUNSD} has played a fundamental role in Document AI, drawing increasing attention from both industry and academia. The task mainly includes the recognition of semantic text entities (also known as \textit{entity labeling} or \textit{named entity recognition}) and the extraction of relationships between them (also referred to as \textit{entity linking}) from Visually-rich Documents (VrDs) like administrative forms and invoices. Language-based DU approaches~\cite{hong2022bros,huang2022layoutlmv3,luo2023geolayoutlm} have proven to be the current state-of-the-art for both tasks. But these approaches have the following drawbacks: 1) They are impractical for deployment in real-world industrial scenarios, where computational resources may be limited, and processing efficiency is vital. 2) They rely heavily on large-scale pretraining to learn upon on a single language (mainly English) making them constrained in terms of their applicability in multilingual scenarios 3) The presence of sensitive content within these business documents often restricts access during the training phase, necessitating the development of DU models capable of learning from only geometrical constraints and the perception of the layout (eg. coordinates of the word bounding boxes). 

Given forms or invoices in an unfamiliar language, humans can generally deduct the composed text entities and their relationships using layout cues and some  experience or prior which they try to approximate. Administrative documents frequently exhibit a semi-structured format, lacking a consistent layout but containing a shared group of elements such as headers, footers, senders, recipients and some entities. This spatial arrangement can often be perceived as a table-like layout. Graphs can effectively capture such topological features of documents with table-like layouts by representing the spatial and hierarchical relationships between document elements in a structured manner. 

Inspired by prior works on Graph Neural Networks (GNNs)~\cite{carbonell2021named,davis2021visual,gemelli2022doc2graph,riba2019table,voutharoja2023language} to interpret administrative documents using mainly the layout information, we propose a two-staged GNN model called \textit{GeoContrastNet} that does not require language information and could be potentially applied to visually similar languages without requiring fine-tuning. Also, existing GNN methods~\cite{davis2021visual,gemelli2022doc2graph} have mainly relied on learning a message passing between the different node components (eg. classes of text segments like header, sender, recipient etc.) to predict the pairwise relationship between the edges (entity linking). Contrary to this, we propose a simple contrastive training strategy on the GNN~\cite{kipf2016semi} to learn some robust edge features that include spatial proximity (how close elements are to each other), hierarchical relationships (parent-child relationships between elements, such as a table and its cells), and the sequential order (the reading order of text blocks). The key intuition is such representative grounded edge features learnt during this contrastive training could essentially serve as a strong prior in the second stage which uses a Graph Attention Network (GAT)~\cite{velivckovic2017graph} to solve both node classification (for entity recognition) and edge classification (for entity labeling) simultaneously. 

Poor quality scans, variations in resolution, or inconsistencies in formatting can often hinder the ability to accurately extract and utilize visual features as observed in Doc2Graph~\cite{gemelli2022doc2graph}. To alleviate this issue, GeoContrastNet introduces a new grounding mechanism that guides the graph attention to combine visual and geometric features. With experimental evidence, we show the utility and effectiveness of visual features when combined with rich representative geometric priors learnt during the contrastive stage. The novelties of this work can be divided into four folds: 1) We propose a two-staged language-agnostic GNN framework, GeoContrastNet, that introduces a simple contrastive learning strategy in the first stage to learn robust and generalized edge features over document samples.  
2) The framework also introduces graph attention in the second stage to ground the previously learnt edge features (representing key-value components) with the visual features. 
3) A comprehensive analysis of the different sets of geometric features (both nodes and edges) has been studied and their utility for the entity recognition and labeling tasks. 4) To justify the effectiveness of the geometric features learnt during training, we also show a quantitative evaluation of our geometric-only model for invoice understanding task.

%% file: tex/related.tex
\section{Related Work}

\noindent
\textbf{Layout Representation Learning.}
The state-of-the-art DU foundation models~\cite{huang2022layoutlmv3,xu2020layoutlm,xu2020layoutlmv2,appalaraju2021docformer,powalski2021going} relies on large-scale pretraining focusing more on language rather than visual or geometrical elements for solving document intelligence tasks like classification~\cite{RVL-CDIP}, information extraction~\cite{FUNSD,huang2019icdar2019} and document visual question answering~\cite{mathew2021docvqa,mathew2022infographicvqa}. They introduced spatial layout information through 2D bounding box coordinates from an OCR as layout features to the language model. Other approaches~\cite{li2021selfdoc,gu2021unidoc} have focused on representing layout at the region level, identifying logical components such as paragraphs, figures, titles and tables which are essential for tasks like document layout analysis~\cite{biswas2021beyond,biswas2022docsegtr,maity2023selfdocseg,banerjee2023swindocsegmenter}. GeoLayoutLM~\cite{luo2023geolayoutlm} adds geometric constraints over LayoutLMv3~\cite{huang2022layoutlmv3} using geometry-based pre-training objectives between the text segments in a self-supervised fashion. This helps them improve significantly on key-value entity linking tasks for forms~\cite{FUNSD}. Motivated by the effectiveness of layout features for several document understanding tasks, we propose a novel contrastive paradigm to learn geometrical layout representation for VrDs.          

\noindent
\textbf{Language-Agnostic Document Understanding.} Most of the DU foundation models are built upon heavy reliance on pretraining with language features to solve downstream tasks. Existing language-agnostic DU models like LayoutXLM~\cite{xu2021layoutxlm} and LILT~\cite{wang2022lilt} also incorporate large-scaled multilingual pre-trained models. Davis \textit{et. al.}~\cite{davis2021visual} addressed this issue by achieving almost the same level of entity linking performance on the FUNSD~\cite{FUNSD} form understanding dataset by learning only visual features from the small provided training data using a Graph Convolutional Network (GCN) backbone. Voutharoja \textit{et. al.}~\cite{voutharoja2023language} used a neuro-symbolic approach that uses an entity-relation graph for scanned forms. Although it achieves great results on the FUNSD benchmark, the features cannot be generalized in a more practical industrial scenario where data is online. Inspired by the graph language models~\cite{gemelli2022doc2graph,wang2023docgraphlm}, we build upon the promising geometric (both node and edge) representation learning which could help us build a language-agnostic DU model that does not utilize OCR but rather focuses on learning key-value relationships in document forms and invoices from a purely visual perspective.

%% file: tex/method.tex
\section{Method}
\begin{figure}[htb]
\includegraphics[width=\textwidth]{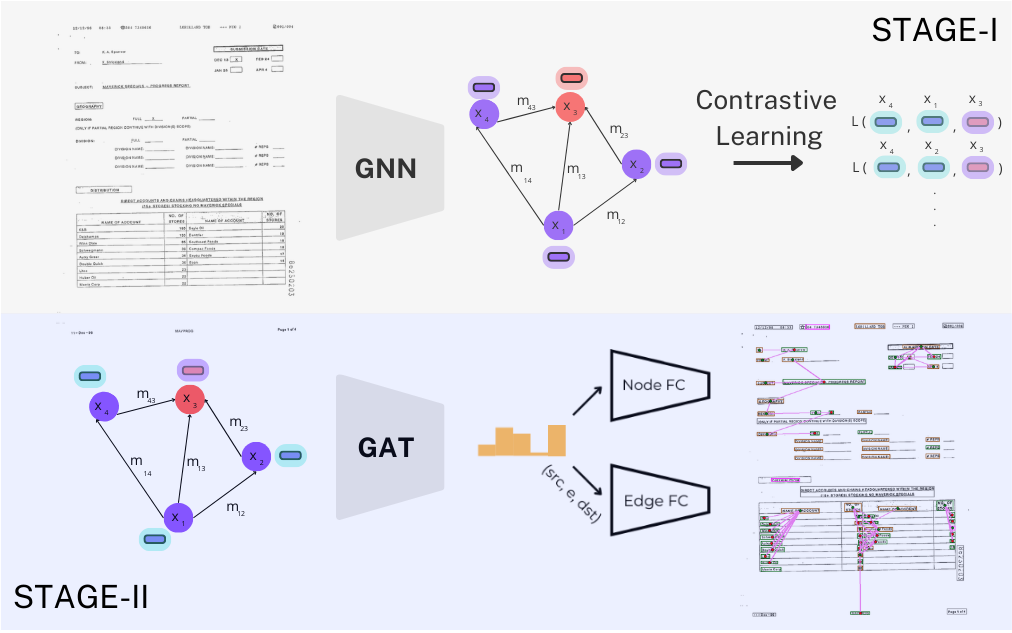}
\caption{\textbf{Overall Architecture of the Proposed GeoContrastNet Framework.}In Stage I, a GNN processes a document image represented as an attributed graph, learning key-value edge features in a contrastive setting using the triplet margin loss. Then, in Stage II, a GAT traverses these features to predict entity labels (nodes) and their relations (edges) as output.} 
\label{fig1}
\end{figure}

In this section, we will look closely into the proposed methodology of GeoContrastNet, the formulation of the problem, the two stages (task modules) incorporated in GeoContrastNet framework, and the learning objectives employed in the architecture shown in Figure \ref{fig1}.

\subsection{Graph-based Document Representation}
Given a document image, it is represented as an attributed graph. The extraction of the document objects of the layout is performed by a layout segmentation process using an off-the-shelf Optical Character Recognition (OCR) provided in the ground truth or using the YOLOv5~\cite{YOLOv5} algorithm to get the bounding box regions of the page text regions. The image is segmented into text regions, represented by the corresponding bounding boxes. Graph nodes correspond to these text bounding boxes, attributed with the geometric attributes: its bounding box, the area of the bounding box, and a regional encoding that encapsulates the node's position within the document. Specifically, the regional encoding is designed to capture the spatial distribution of nodes, providing a comprehensive representation of each node's context. Graph edges represent the structural information of the document. The links between the nodes are constructed using the $k$-NN algorithm, where each node is linked with its $k$ nearest neighbours in terms of spatial information. Each edge \(m_{ij}\) that connects node \(i\) to node \(j\), is represented by a feature vector that includes several geometric attributes: the angle \(\theta_{ji}\) between nodes \(i\) and \(j\), measured in a standard reference frame; the Euclidean distance \(d_{ij}\) between the nodes, which reflects their spatial separation; discretized polar coordinates that offer a granular view of the relative positioning; and the relative positions of \(i\) and \(j\), which provides information for understanding the directional relationships between nodes.

\subsection{Method Overview}
\begin{figure}[htb]
\includegraphics[width=\textwidth]{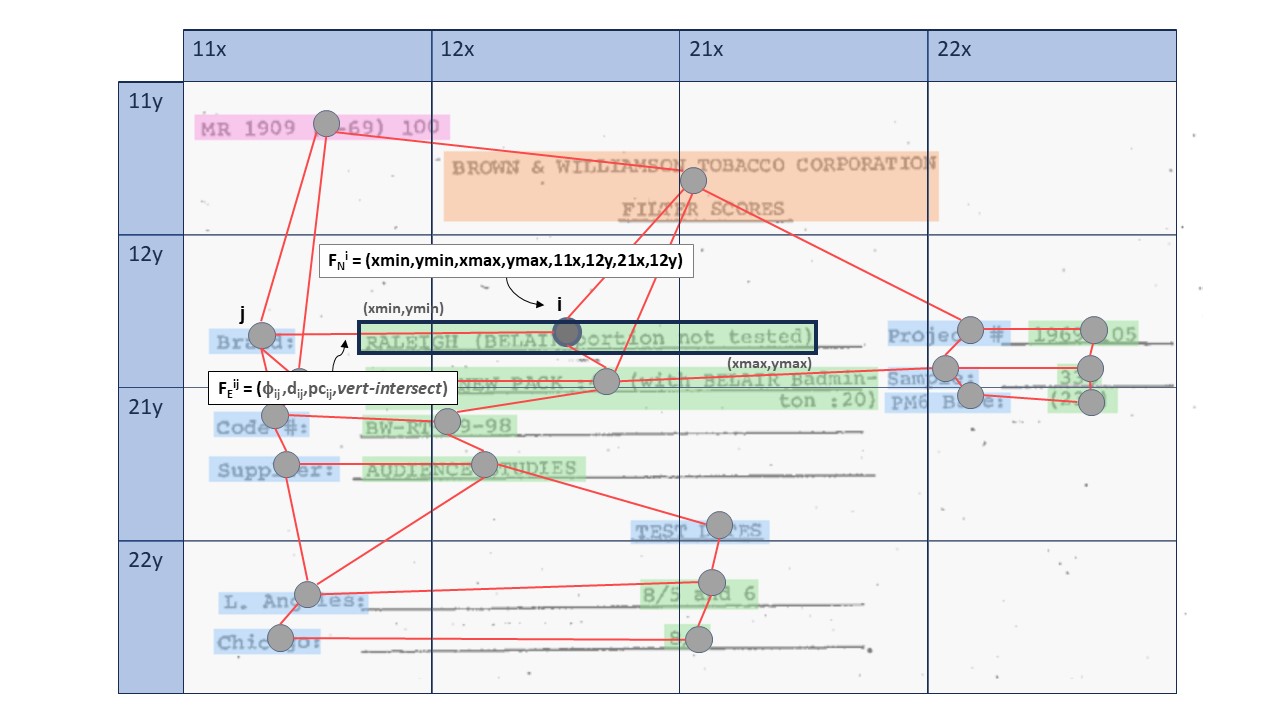}
\caption{\textbf{Visual Illustration of Proposed Node and Edge Representation}. Node features are defined by the bounding box coordinates, the area and regional encoding providing local and global structure position. Edge features provide information to geolocate each node relative to its neighbors.} 
\label{Graph}
\end{figure}

A document image, after segmenting the text regions is represented as an attributed graph $G(N,E,F_N,F_E)$ where $N$ is the set of nodes, $E$ is the set of edges or links between nodes, and $F_N$ and $F_E$ are the attributes or features vectors characterizing respectively the nodes and the edges. These feature vectors are defined as follows (see Fig. \ref{Graph} for a graphical illustration): 

\noindent
\textbf{Node Representation}. Given a node \( i \), it is represented by the feature vector \( F_N^i = (xmin, ymin, xmax, ymax, a, r_{xmin}, r_{ymin}, r_{xmax}, r_{ymax}) \), where:
\begin{itemize}
    \item The bounding box is represented by the coordinates \textit{xmin}, \textit{ymin}, \textit{xmax} and \textit{ymax}, defining its position and size.
    \item The area $a$ of the bounding box.
    \item For each bounding box coordinate \( xmin \), \( ymin \), \( xmax \), and \( ymax \), a regional encoding is defined as \( r_{xmin} \), \( r_{ymin} \), \( r_{xmax} \), and \( r_{ymax} \). These encodings are derived from the normalize coordinates relative to the size of the image, with the largest dimension adjusted to a scale of 1. After this normalization, the document is conceptually divided into four equal sections along a single dimension. A code from the set \{\textit{11}, \textit{12}, \textit{21}, \textit{22}\} is assigned to each normalized coordinate based on which of these sections it falls into. This approach ensures that every coordinate is consistently and accurately represented within the normalized structure of the document.

\end{itemize}
{\bf Edge Representation}. Given two nodes $i,j$, an edge $(i,j)$ that links them is represented by the feature vector \(F_E^{ij}=(\theta_{ji},d_{ij},pc_{ij}, rp_{ij})\), where:
\begin{itemize}
    \item The angle \(\theta_{ji}\) represents the orientation of the source node relative to the destination node, encapsulating the directional relationship between them.
    \item The Euclidean distance $d_{ij}$ measures the direct spatial separation between two nodes, providing a quantitative assessment of their proximity.
    \item The relative positioning of nodes is determined using discretized polar coordinates $pc_{ij}$. Each source node is positioned at the center of a Cartesian plane, facilitating the encoding of its neighbors based on distance and angle relative to it. The space is divided into various bins (one-hot encoded), allowing for a selectable number of partitions.
    \item Relative positions $rp_{ij}$ of nodes to provide more information of a global position of the nodes inside the document. The position is encoded based in a dictionary of language tokens that describe relative positional information: {\em left, right, top, bottom, vert-intersect, hor-intersect,sqr-intersect}.
\end{itemize}

Documents have a rich geometrical information that play a crucial role in understanding the overall structure of the entities. We propose two independent stages to address the geometric edge feature learning and the prediction phase.
\begin{itemize}

\item In \textit{Stage-I}, during this phase, we explore the broad applications of the Graph Neural Network (GNN) model, focusing on its ability to process and learn from node and edge information. Key concepts in this phase include the implementation of triplet loss and a contrastive setting, which are instrumental in enhancing the model's understanding of the graph's topology and the relationships between its elements.

\item In \textit{Stage-II}, the model's learning objectives expand to encompass the joint learning of semantic entities and link prediction. We finetune a Unet to extract visually rich features, that then are combined with the features obtain in \textit{Stage --I} to produce the input vector for the GAT layers.
\end{itemize}

In the following subsections we further describe the mentioned stages.

\subsection{Stage -- I: Geometric Edge Feature Learning}

GNNs are neural models designed to capture dependencies within the data that is represented in the graph nodes and edges. This is achieved by the concept of {\em message passing} consisting in propagating information between nodes and aggregating this information in node embeddings. In this way, at each layer $n$ , each node is capturing the information of nodes that are $n$ hops away, so the context is encoded in the node embedding. The Stage-1 architecture includes a GNN with a modified aggregation function to better process the geometric information from the edges and the nodes. 

Graph Contrastive Learning (GCL) learns representations by contrasting positive samples against negative ones. Positive samples are similar graph nodes or edges, while negative samples are dissimilar ones. By maximizing the agreement between representations, GCL allows us to learn meaningful graph features. The GNN is trained using a triplet margin loss function,
aiming to refine the representation of node geometric information by leveraging the geometric data derived from the edges. The \textit{Message Formation} for each node $u_i$ belonging to the set of nodes in the graph, involves the outgoing message $m_{ij}$ $i \neq j$, being determined by the feature information of the edge, such that $m_{ij} = e_{ij}$.


The outgoing node aggregates the representation of the edge features of its immediate neighborhood, \{$h_v, \forall v \in \mathcal{N }(u)$\}. The aggregation is defined by:
\begin{equation}
    h'_v = \frac{c}{|\Upupsilon(i)|} \sum_{j \in \Upupsilon(i)}m_{ij}
\end{equation}where $\Upupsilon(i) = \{ v \in \mathcal{N}(u) : \|u - v\| < \text{threshold} \}$, $\|u - j\|$ is the Euclidean distance of nodes $u$ and $v$ normalized between 0 and 1 and $c$ is a constant scale factor.
Following the aggregation process, the updated feature vector  \(F^{i+1}_N\) is formed by concatenating the outgoing node's feature vector \( F_N^i \) with the newly aggregated value \( h'_v \), this concatenated vector is then used as input to a linear transformation, normalized by a layer normalization and subsequently passed through a ReLU activation function to yield the updated node representation.
\begin{equation}
    F^{i+1}_N(i)  = \text{ReLU}\left(\text{LayerNorm}\left(\textbf{W} \cdot [F_N^i \; \| \; h'_v]\right)\right)
\end{equation}

where, \(\textbf{W}\) is the weight matrix associated with the linear transformation, \([F_N^i\; \| \; h'_v]\) denotes the concatenation of \(F_N^i\) and the aggregation result, and \(\text{ReLU}\) and \(\text{LayerNorm}\) represent the ReLU activation function and layer normalization, respectively. After the GNN, the new node representations based on geometric features are used for calculating the contrastive loss.

\subsection{Stage -- II: Prediction Phase}
The design for the second stage employs a two layer Graph Attention Network (GAT).  GATs leverages attention layers to allow nodes to attend over their neighborhood features. By specifying different weights to different nodes in a neighborhood, GATs enhance expressiveness without prior knowledge of the graph structure. Our GAT receives enhanced geometric edge representations from the initial phase, integrating them with visual embeddings. The source of these visual embeddings is a U-net encoder, that is trained while performing both entity linking and semantic entity labeling tasks.

\subsubsection{Semantic Entity Labeling.}
The output from the Graph Attention Network (GAT) layer, denoted as \( h \), is subsequently processed through a sequence of five linear projection layers. These layers transform \( h \) to match the dimensionality corresponding to the desired number of prediction classes. The transformation is concluded with the application of a softmax function to obtain probability distributions over the classes, followed by the argmax operator to determine the predicted entity for each node.

\subsubsection{Entity Linking.}
We adopt the edge representation as proposed in Doc2Graph~\cite{gemelli2022doc2graph}. Each edge is treated as a triplet structure \( (src, e, dst) \) where the edge representation \( h_e \) is defined by:
\begin{equation}
    h_e = h_{src} \parallel h_{dst} \parallel cls_{src} \parallel cls_{dst} \parallel e_{polar}
\end{equation}
    Here, \( h_{src} \) and \( h_{dst} \) are sourced from the node embeddings produced by the final layer of our Graph Attention Network (GAT). The softmax probabilities, \( cls_{src} \) and \( cls_{dst} \), come from the node prediction layer's output logits, and \( e_{polar} \) refers to the polar coordinates as outlined in the original paper's Section 3.2. We note as well as in Doc2Graph that using the class information is helpful for the model to predict if there exists a link between two types of nodes. Finally \( h_e \) is fed into five layers of the Fully-Connected (FC) classifier.

\subsection{Learning Objectives}
In this subsection, we discuss the key objective functions used in our model training for both Stage-I and Stage-II. 

\noindent
\textbf{Stage -- I Objectives.}
\noindent In this stage the learning objective is to construct a new geometric representation for each of the N entities present in the graph. To do so we use a \textit{contrastive learning loss}, more specifically a triplet margin loss, defined as follows:

\begin{equation}
L(a, p, n) = \max\{d(a_i, p_i) - d(a_i, n_i) + \text{margin}, 0\}
\end{equation}
where
\begin{equation}
d(x_i, y_i) = \|x_i - y_i\|_p
\end{equation}

\noindent
\textbf{Stage -- II Objectives.} During this phase, the Graph Attention Network (GAT) and the Unet encoder were trained jointly on both objectives of link prediction and semantic entity recognition. The training utilized a cross-entropy loss function for both objectives, with the final loss being a summation of the individual losses from each task. The overall loss objective can be mathematically represented as:

\begin{equation}
\mathcal{L}_{\text{total}} = \mathcal{L}_{\text{entity}} + \mathcal{L}_{\text{link}}
\end{equation}

where \(\mathcal{L}_{\text{entity}}\) and \(\mathcal{L}_{\text{link}}\), are described by the following:

\begin{equation}
\mathcal{L}_{\text{entity}} = - \sum_{i=1}^{N} y_{\text{entity}, i} \log(\hat{y}_{\text{entity}, i})
\end{equation}

\begin{equation}
\mathcal{L}_{\text{link}} = - \sum_{j=1}^{E} y_{\text{link}, j} \log(\hat{y}_{\text{link}, j})
\end{equation}

Here, \(N\) and \(E\) signify the total number of entities and links, respectively. The symbols \(y_{\text{entity}, i}\) and \(y_{\text{link}, j}\) correspond to the actual labels for entities and links, while \(\hat{y}_{\text{entity}, i}\) and \(\hat{y}_{\text{link}, j}\) denote the predicted probabilities for the entities and links, respectively.

%% file: tex/experiments.tex
\newcommand{\cmark}{\ding{51}}
\newcommand{\xmark}{\ding{55}}

\section{Experimental Validation}
In this section, we introduce the experimental validation of our proposed method. Finally, we show a series of ablation studies to justify the effectiveness of the different components in our model. 

\subsection{Dataset Description}
We evaluate our method on two datasets, the FUNSD dataset and the RVL-CDIP Invoices. The FUNSD dataset consists of 199 authentic, fully annotated scanned forms. These documents are a curated selection from the broader RVL-CDIP~\cite{RVL-CDIP} collection, which contains 400,000 grayscale images of diverse documents. The overall dataset is organized into a training set with 149 samples and a test set comprising 50 samples. For model validation, we employ a random partitioning strategy on the training set to create a validation subset. Our evaluation covers semantic entity recognition, categorizing entities into "question," "answer," "header," or "other," as well as link prediction tasks. In the work of Riba \textit{et.al.}~\cite{riba2019table} another subset of RVL-CDIP has been released. The authors selected 518 documents from the invoices classes, annotating 6 different regions, namely: "invoice info", "other", "positions", "receiver", "supplier", "total". The task that can be performed are layout analysis, in terms of node classification, and table detection, in terms of bounding box IoU threshold greater than 0.5.

\subsection{Evaluation Metrics}

We present our findings across two main tasks: link prediction and semantic entity recognition for FUNSD. In the context of link prediction, we evaluate our model using three metrics: F1 score for non-entity links (F1 None), F1 score for key-value pairs (F1 Key-Value), and the Area Under the Curve (AUC). For semantic entity recognition, we focus on the micro-averaged F1 score (F1 Micro) to assess overall performance. For RVL-CDIP Invoices we evaluate on table detection and on layout analysis.

\subsection{Implementation details}

\subsubsection{Stage-I} consists of a two-layer Graph Convolutional Network (GCN). The initial layer begins with a 9-dimensional node vector, which is merged with a 15-dimensional vector from message passing to form a 24-dimensional vector. This vector is processed through an MLP, followed by layer normalization and a ReLU activation, resulting in a 15-dimensional vector. The second layer adopts a similar approach, where it projects a 30-dimensional vector through an MLP, reducing it to 17 dimensions.

\subsubsection{Stage-II} incorporates the learned representations from Stage-I along with visual features extracted from each bounding box using a MobileNet encoder~\cite{howard2017mobilenets} from a pretrained UNet. The first GAT layer projects dimensions from 1465 to 1500. The subsequent layer includes two heads for multi-head attention, expanding the dimensions from 1500 to 3000. To prevent overfitting, each GAT layer incorporates residual connections and applies a 20\% dropout to both the features in the attention mechanism and the attention weights. For downstream tasks, two modules handle entity linking and semantic entity labeling, respectively. Each module takes the output features from the GAT and maps them to the respective number of classes required for each task. For semantic entity labeling, five MLPs project from 3000 to 4 classes. In the entity linking task, five MLPs are used to map from 6014 to 2 classes.

\subsubsection{Hardware}
We trained GeoContrastNet using a single NVIDIA GeForce RTX 3090. The entire training process lasts 1 hour and 10 minutes overall, with the Stage-I phase taking only 10 minutes and Stage-II taking the rest 1 hour.

\begin{table}[t]
\centering
\setlength\tabcolsep{5pt} 
\resizebox{\textwidth}{!}{
\begin{tabular}{lllllll}
\hline
 & & & \multicolumn{2}{c}{\footnotesize{\(F_1\)} (\(\uparrow\))} &  \\
 \cline{4-5}
 \addlinespace[0.5em]
\footnotesize\textbf{Method} & \footnotesize\textbf{Modalities} &\footnotesize\textbf{GNN} & \footnotesize\textbf{SER} & \footnotesize\textbf{EL} & \footnotesize\textbf{\# Params \(\times10^6\)} \\
\addlinespace[0.3em]
\hline
BROS ~\cite{hong2022bros} &          T + V    & \(\times\) & 0.8121 & 0.6696 & 138 \\
LayoutLM~\cite{xu2020layoutlm} &   T + V         & \(\times\) & 0.7895 & 0.4281 & 343 \\
                &                 &           &        &         &\\
FUNSD~\cite{FUNSD}      &       T + G         & \(\checkmark\) & 0.5700 & 0.0400 & - \\
FUDGE~\cite{davis2021visual} &    V + G        & \(\checkmark\) & 0.6507 & 0.5241 & 12 \\
Doc2Graph~\cite{gemelli2022doc2graph}  &     T + G + V     & \(\checkmark\)  & 0.8225 & 0.5336  & 6.2 \\
Voutharoja et. al.~\cite{voutharoja2023language} &      G       &   \(\checkmark\)   &  0.8225     & 0.8540 & 0.0000081    \\
&  & & & & \\
GeoContrastNet(Ours) + YOLO       &   G + V    & \(\checkmark\) & {0.5260} & {0.2438} & 14 \\
GeoContrastNet(Ours) + GT         &   G + V    & \(\checkmark\) & \textbf{0.6476} & \textbf{0.3245} & 14 \\
\hline
\end{tabular}
}
\newline
\caption{\textbf{SOTA Comparison on FUNSD}. The results have been shown for both semantic entity labeling (SEL) and entity linking (EL) tasks with their corresponding metrics where T:Text, G:Geometry, V:Visual}
\label{table:funsd_sota}
\end{table}

\subsection{Competitors}
As shown in Table~\ref{table:funsd_sota}, we show a fair comparison of our proposed method with the existing SOTA. The results show that we achieve promising results on the semantic entity labeling task with 64.76\% F1 score among language-agnostic approaches which is almost on par with FUDGE~\cite{davis2021visual}. For the entity linking task, we achieve a decent score of 32.45\%, although it lags a bit behind our competitors Doc2Graph~\cite{gemelli2022doc2graph} and FUDGE~\cite{davis2021visual}. While Doc2Graph~\cite{gemelli2022doc2graph} uses the text supervisory signal massively to boost the performance of it's model. On the other hand, FUDGE~\cite{davis2021visual} largely performs better mainly due to the effective interplay between the geometric and visual features in their GCN architecture pipeline. Although GeoContrastNet learns rich geometric representation in Stage-I, the simple feature concatenation adapted in Stage-II doesn't give it a huge boost when compared with FUDGE. On the other hand, Voutharaja \textit{et. al.}~\cite{voutharoja2023language} is not a robust end-to-end differentiable approach as they construct a heuristical entity-relation graph with some heavily handcrafted priors to train their GNN for FUNSD. This gives them extremely high performance on the linking task but it's not designed for generalizability in a real-world practical setting.

\begin{table}[ht]
\centering
\resizebox{\textwidth}{!}{%
\setlength{\tabcolsep}{5pt} 
\begin{tabular}{@{}cccccccccccc@{}}
\toprule
\multicolumn{4}{c}{\footnotesize{Edge Features}} & & \multicolumn{3}{c}{\footnotesize{Node Features}} & & \multicolumn{2}{c}{\footnotesize{\(F_{1}\) per classes 
 \((\uparrow)\)}} &  \\
\cline{1-4} \cline{6-8} \cline{10-11}
\addlinespace[0.5em]
\footnotesize\textbf{Distance} & \footnotesize\textbf{Angle} & \footnotesize\textbf{Discretize} & \footnotesize\textbf{Polar} & & \footnotesize\textbf{Bounding} & \footnotesize\textbf{Area} & \footnotesize\textbf{Regional} & \footnotesize\textbf{\textbf{\(F_1\)} Nodes} (\(\uparrow)\) & \footnotesize\textbf{None} & \footnotesize\textbf{K-V} & \footnotesize\textbf{AUC-PR} (\(\uparrow)\) \\
\midrule
\cmark & \cmark & \cmark & \cmark & & \cmark & \cmark & \cmark & 0.5049 & 0.8933 & 0.2017 & 0.5737 \\
\xmark & \cmark & \cmark & \cmark & & \cmark & \cmark & \cmark & 0.4366 & 0.8825 & 0.1931 & 0.5767 \\
\cmark & \xmark & \cmark & \cmark & & \cmark & \cmark & \cmark & 0.4051 & 0.8585 & 0.1808 & 0.5724 \\
\cmark & \cmark & \xmark & \cmark & & \cmark & \cmark & \cmark & 0.5062 & 0.8805 & 0.1898 & 0.5761 \\
\cmark & \cmark & \cmark & \xmark & & \cmark & \cmark & \cmark & 0.5244 & 0.8923 & 0.2059 & 0.5857 \\
\cmark & \cmark & \cmark & \cmark & & \xmark & \cmark & \cmark & 0.5624 & 0.8779 & 0.1868 & 0.5850 \\
\cmark & \cmark & \cmark & \cmark & & \cmark & \xmark & \cmark & 0.5412 & 0.9044 & 0.2168 & 0.5989 \\
\cmark & \cmark & \cmark & \cmark & & \cmark & \cmark & \xmark & \textbf{0.5750} & \textbf{0.9120} & \textbf{0.2290} & \textbf{0.6104} \\
\bottomrule
\end{tabular}
}
\newline
\caption{\textbf{Ablation on Geometric Features}. We report results for different combinations of geometric features used in Stage-I during message passing.}
\label{tab:ablation_study_features}
\end{table}

\subsection{Ablation Studies}
We conducted extensive ablation studies to analyze the efficiency and generalizability of our method components.
A series of tests were carried out using the GeoContrastNet framework. 

\noindent
\textbf{Effectiveness of Geometric Features in Stage-I:}
To understand how various geometric features influence our results in the message passing in Stage-I, we examine the contribution of features derived from edges, nodes, or a combination of both, as detailed in Table~\ref{tab:ablation_study_features}. Our baseline incorporates all geometric features from both edges and nodes in the message passing, and we assess how each specific feature affects the performance in both entity linking and semantic entity recognition tasks. The best geometric features obtained in Stage-I correspond to the last line in Table~\ref{tab:ablation_study_features}. We also report results from using only the geometric features of the edges or the nodes in Table~\ref{tab:ablation_study}, where we observe that edge geometric information alone gives better results on both tasks compared to node geometric information. We hypothesize that edge information contains more spatial information of the surroundings, compared to node geometric information, enabling the model to perform better on both tasks. 


\begin{table}[ht]
\centering
\resizebox{\textwidth}{!}{%
\setlength{\tabcolsep}{8pt} 
\begin{tabular}{@{}llllllllllll@{}}
\toprule
\multicolumn{4}{c}{\footnotesize{Edge Features}} & & \multicolumn{3}{c}{\footnotesize{Node Features}} & & \multicolumn{2}{c}{\footnotesize{\(F_{1}\) per classes \((\uparrow)\)}} &  \\
\cline{1-4} \cline{6-8} \cline{10-11}
\addlinespace[0.5em]
\footnotesize\textbf{Distance} & \footnotesize\textbf{Angle} & \footnotesize\textbf{Discretize} & \footnotesize\textbf{Polar} & & \footnotesize\textbf{Bounding} & \footnotesize\textbf{Area} & \footnotesize\textbf{Regional} & \footnotesize\textbf{\textbf{\(F_{1}\)} Nodes} \((\uparrow)\) & \footnotesize\textbf{None} & \footnotesize\textbf{K-V} & \footnotesize\textbf{AUC-PR} \((\uparrow)\) \\
\midrule
\cmark & \cmark & \cmark & \cmark & & \xmark & \xmark & \xmark & \textbf{0.5313} & \textbf{0.9056} & \textbf{0.2264} & \textbf{0.5871} \\
\xmark & \xmark & \xmark & \xmark & & \cmark & \cmark & \cmark & 0.3532 & 0.8635 & 0.1798 & 0.5579 \\
\bottomrule
\end{tabular}
}
\newline
\caption{\textbf{Node vs Edge Features}. Results on both tasks when using either edge or node geometric information}
\label{tab:ablation_study}
\end{table}

\begin{table}[ht]
\centering
\resizebox{\textwidth}{!}{%
\setlength{\tabcolsep}{8pt} 
\begin{tabular}{@{}lllll@{}}
\toprule
\multicolumn{2}{c}{\footnotesize{Features}} & & \multicolumn{2}{c}{\footnotesize{\(F_{1}\) per classes \((\uparrow)\)}}\\
\cline{1-2}  \cline{4-5}
\addlinespace[0.5em]
\footnotesize\textbf{Stage-I Geometric} & \footnotesize\textbf{Visual} & \footnotesize\textbf{AUC-PR} \((\uparrow)\) & \footnotesize\textbf{None} & \footnotesize\textbf{Key-Value} \\
\midrule
\xmark & \cmark & 0.5483 & 0.8314 & 0.1581 \\
\cmark & \xmark & 0.5871 & 0.9056 & 0.2264 \\
\cmark & \cmark & \textbf{0.6375} & \textbf{0.9825} & \textbf{0.3245} \\
\bottomrule
\end{tabular}
}
\newline
\caption{\textbf{Ablation Study for Modalities}. Results of the different combination of modalities in GeoContrastNet.}
\label{tab:ablation_study_modality}
\end{table}

\begin{figure}[h]
    \centering 
    \begin{minipage}[h]{0.5\linewidth} 
        \includegraphics[width=7cm]{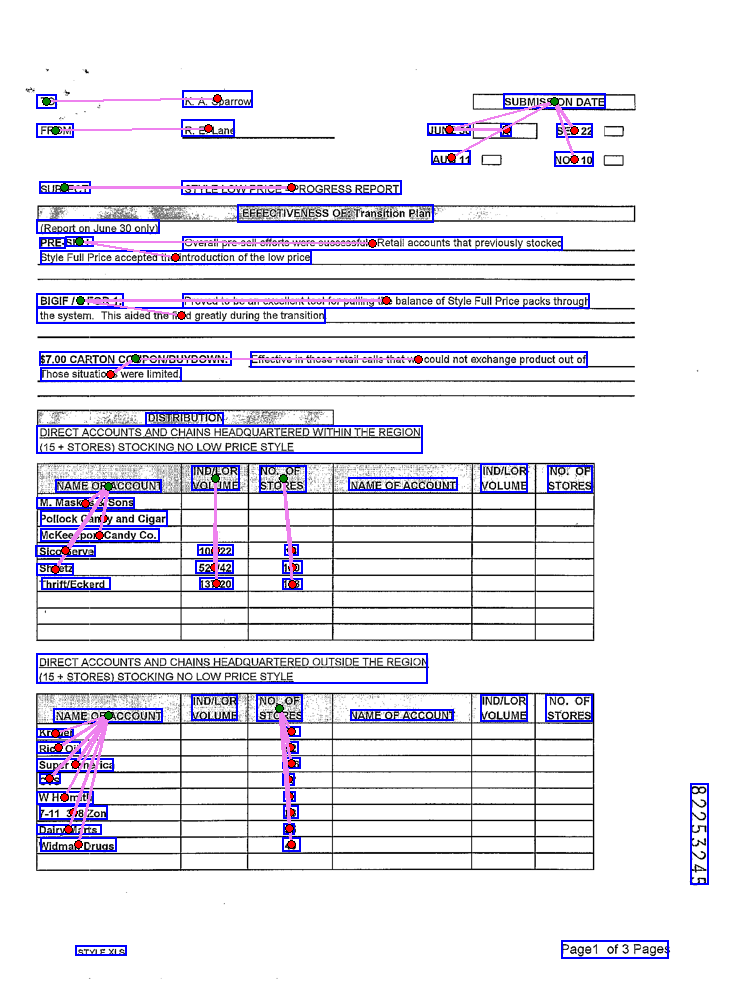}
    \end{minipage}%
    \begin{minipage}[h]{0.5\linewidth}
        \includegraphics[width=7cm]{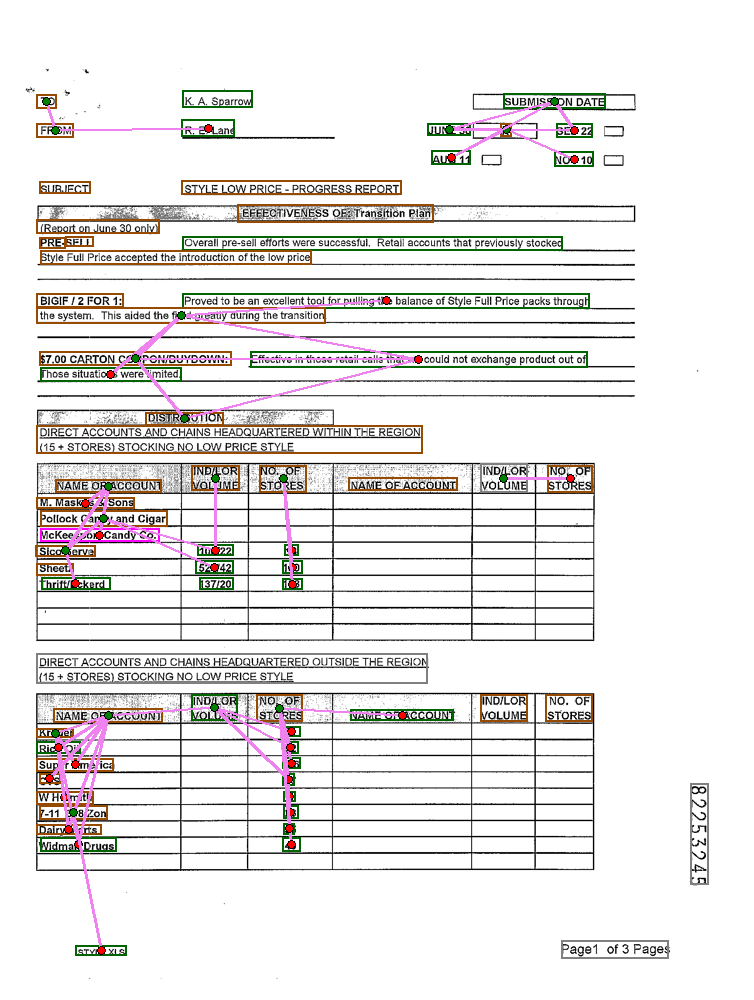}
    \end{minipage}
    \caption{Prediction on the link prediction task on \textbf{FUNSD dataset.} From (L to R) GT and predicted images respectively.}
    \label{fig:FUNSD}
    \vspace{-10pt} 
\end{figure}

\begin{table}[ht]
\centering
\setlength{\tabcolsep}{8pt}
\begin{tabular}{@{}lllll@{}}
\toprule
& & \multicolumn{3}{c}{Metrics \((\uparrow)\)}\\
\cline{3-5}
\addlinespace[0.5em]
\textbf{Method} & \textbf{Threshold} & \textbf{Precision} & \textbf{Recall} & \textbf{\(F_1\)} \\
\midrule
Riba et al.~\cite{riba2019table} & 0.5 & 0.1520 & 0.3650  & 0.2150\\
GeoContrastNet(Ours)             & 0.5 & \textbf{0.2718}  & \textbf{0.2669}&\textbf{0.2693}\\
\bottomrule
\end{tabular}
\newline
\caption{\textbf{Table Detection in terms of F1 score}. A table is considered correctly detected if its IoU is greater than 0.50. Threshold values refers to edges to not be cut: in our case is set to 0.50 by the softmax in use.}
\label{tab:ablation_study_table}
\end{table}
\subsubsection{Role of Different Modalities:}
In Table~\ref{tab:ablation_study_modality} we evaluate the effect of geometric features with combination with the visual features in the FUNSD dataset. We observe an incredible increment in \(F_1\) scores thanks to the fusion of geometric and visual features.

\subsubsection{RVL-CDIP} We evaluate the proposed two stage model in the RVL-CDIP Invoices dataset. Our model outperforms in layout analysis and table detection as shown in Table~\ref{tab:ablation_study_table}. In particular, for table detection, we extracted the subgraph induced by the edge classified as ‘table’ (two nodes are linked if they are in the same table) to extract the target region. Riba et al.~\cite{riba2019table} formulated the problem as a binary classification: we report, for brevity, in Table~\ref{tab:ablation_study_table} the threshold on confidence score they use to cut out edges, that in our multi-class setting (‘none’ or ‘table’) is implicitly set to 0.50 by the softmax.

\begin{figure}[h!]
    \begin{minipage}[b]{0.5\linewidth}
        \centering
        \includegraphics[width=6cm]{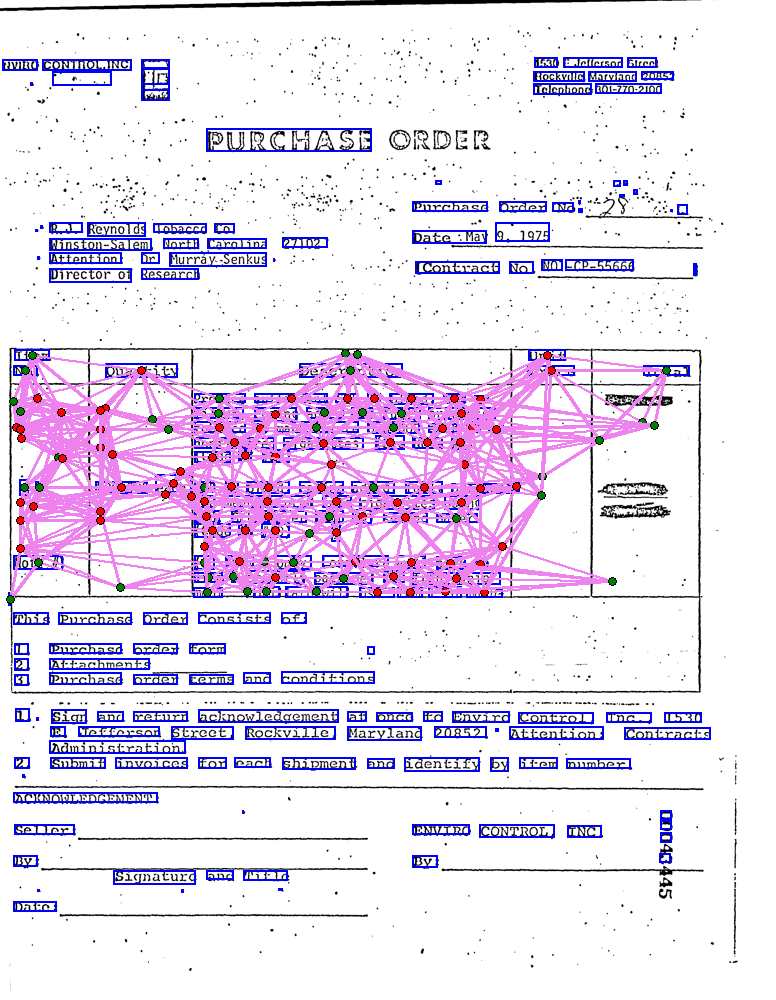}
    \end{minipage}%
    \begin{minipage}[b]{0.5\linewidth}
        \centering
        \includegraphics[width=6cm]{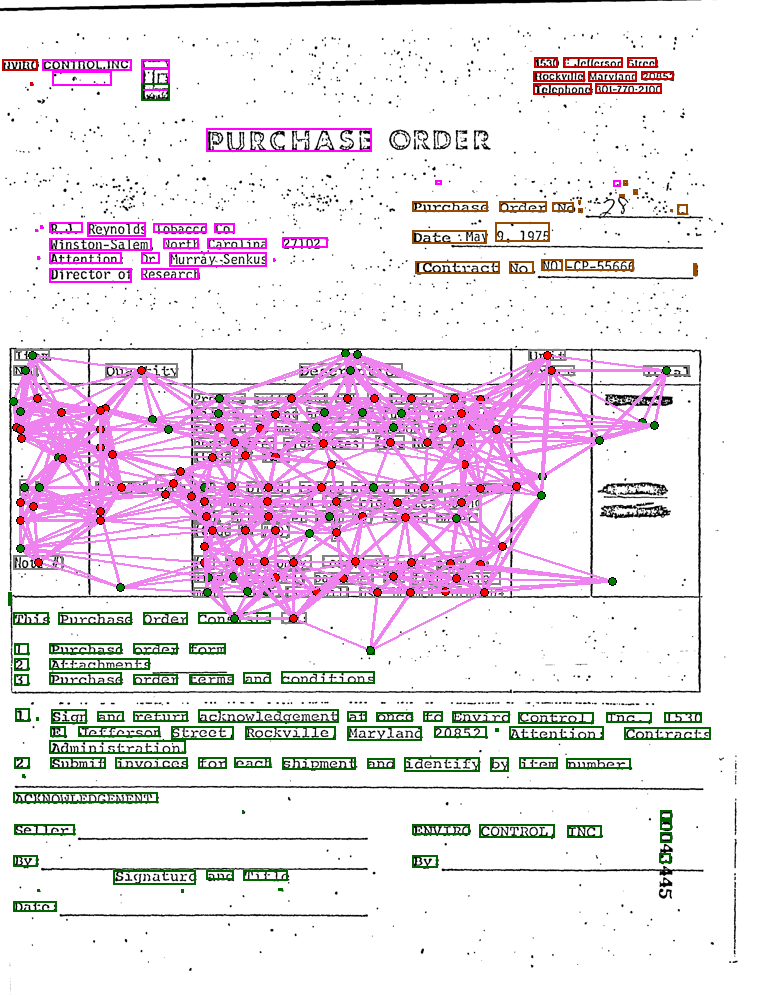}
    \end{minipage}
    \caption{Prediction on the link prediction task on \textbf{RVL-CDIP Invoices.} From (L to R) GT and predicted images respectively.}
    \label{fig:RVL-CDIP}
\end{figure}

\subsection{Discussions and Analysis}
\medtit{Qualitative Discussion on FUNSD} As shown in Figure~\ref{fig:FUNSD} we see an example of a  visually-rich form with ground-truth image compared with the predicted image from our model. The ground truth image (on the left) serves as a benchmark outlining the intended relationships and connections between semantic text entities within the document. It showcases the ideal mapping of links, providing a clear standard against which the predicted outcomes can be evaluated. On the other hand, the predicted image (on the right) our model's attempt to predict the links between the entities, to assess its capabilities. On closely examining, we see that there is a reason why our model gets a very high recall on the edges since it tries to predict a lot of relations in the page. There are some relations (which are not perfectly predicted) and hence it has a lower precision comparatively.

\medtit{Qualitative Discussion on RVL-CDIP Invoices} As shown in Figure~\ref{fig:RVL-CDIP}, we see how the model is able to predict tables correctly. This also shows that the edge classifier performs really well and according to Table~\ref{tab:ablation_study_table} attributes the very high accuracy numbers achieved for the layout analysis task. The much improved precision rate of the model compared to Riba \textit{et. al.}~\cite{riba2019table} is attributed to the geometric edge feature learning phase which results in more correctly predicted key-value links.

\medtit{Information Extraction for Privacy-Preserving} In recent times, we have seen that Document AI is moving towards solving DU tasks on documents that contain sensitive or copyrighted content. A recent challenge has been launched for visual question answering~\cite{tito2023privacy} that deals in such scenario. Our geometric-only model (without integrating any visual or textual modality) contributes a step towards moving in this direction. We show results illustrated in Table~\ref{tab:ablation_study_modality} to actually show the potential of GeoContrastNet for the challenging table detection task without incoroporating any visual or textual information. 

%% file: tex/conclusion.tex
\section{Conclusion and Future Work}

In conclusion, GeoContrastNet offer a powerful and versatile framework for capturing the fine-grained topological features of documents with table-like layouts. By representing document text entities as nodes and explicitly learning their relationships (edge features) using a contrastive strategy, these models show highly promising results on both form and invoice understanding tasks. Also, we study the effectiveness of the visual features in this work by showing how the graph attention module help to align the layout structure with visual components of the page. This two-stage approach highlights the importance of geometric information inside structured documents, which can be beneficial to process complex document layouts in a privacy-preserving document understanding (eg. visual question answering~\cite{tito2023privacy}) setting. 

\noindent
\textbf{Future Scopes and Challenges}
The graph construction, particularly through the use of the K-Nearest Neighbor (KNN) algorithm, presents certain limitations by conditioning both the structure of the graph and the information flow among nodes. The KNN method sets node connections using a fixed constant K, which conditions message transmission by determining the information available to each node at any given moment T. The choice of K can introduce biases that potentially affect the model's outcomes. GeoContrastNet, which employs the KNN algorithm to establish the graph's connections, may inadvertently incorporate these biases, impacting the learning process. Another limitation of GeoContrastNet is the fusion mechanism implemented which can be vastly improved using an attention mechanism~\cite{bahdanau2014neural} to align the vector spaces across the modalities. Also, we would like to explore the ability of our language-agnostic GNN model in multilingual settings~\cite{xu2022xfund} for future work.